\pdfoutput=1
\documentclass[11pt, a4paper, logo, copyright, nonumbering]{hku}
\usepackage{natbib}

\usepackage{fancyhdr}
\usepackage[utf8]{inputenc} 
\usepackage[T1]{fontenc}    
\usepackage{booktabs}       
\usepackage{amsfonts}       
\usepackage{nicefrac}       
\usepackage{microtype}      
\usepackage{xcolor}         

\usepackage{graphicx}
\usepackage{multirow}
\usepackage{enumitem}
\usepackage{subcaption}
\usepackage{xspace}
\usepackage{makecell}
\usepackage{tcolorbox}

\usepackage{fontawesome5}  
\usepackage{array}
\definecolor{mydarkblue}{rgb}{0,0.08,0.45}
\definecolor{linkcolor}{rgb}{0.956,0.298,0.235}
\definecolor{citecolor}{HTML}{1976D2}
\hypersetup{
    colorlinks=true,
    linkcolor=citecolor,
    filecolor=magenta,
    urlcolor=cyan,
    citecolor=citecolor,
}

\usepackage{rotating}
\usepackage{enumitem}
\usepackage[capitalize,noabbrev]{cleveref}
\crefname{section}{Sec.}{Secs.}
\Crefname{section}{Section}{Sections}
\Crefname{table}{Table}{Tables}
\crefname{table}{Tab.}{Tabs.}

\newcommand{\algoName}{HVMDx\xspace}

\definecolor{deemph}{gray}{0.6}

\reportnumber{001}

\renewcommand{\today}{}

\title{\centering Diagnosing Shoulder Disorders Using Multimodal Large Language Models and Consumer-Grade Cameras}
\author[*]{
\small
Jindong Hong$^{1,2,\dag}$ \quad Wencheng Zhang$^{1}$ \quad Shiqin Qiao$^{1}$ \quad Jianhai Chen$^{3}$ \quad Jianing Qiu$^{4}$ 

Chuanyang Zheng$^{5}$ \quad Qian Xu$^{1}$ \quad Yun Ji$^{1}$ \quad Qianyue Wen$^{1}$ \quad Weiwei Sun$^{1}$ \quad Hao Li$^{1}$ 

Huizhen Li$^{1}$ \quad Huichao Wang$^{1}$ \quad Kai Wu$^{1}$ \quad Meng Li$^{1}$ \quad Yijun He$^{1}$ \quad Lingjie Luo$^{1}$ \quad  Jiankai Sun$^{1,\dag}$ \\

\small
$^1$Bytedance \quad $^2$Peking University \quad $^3$Peking University People’s Hospital 

$^4$Mohamed bin Zayed University of Artificial Intelligence \quad $^5$The Chinese University of Hong Kong \\
\quad \\
\small
$^\dag$Corresponding Authors \\
}

\begin{abstract}
Shoulder disorders, such as frozen shoulder (a.k.a., adhesive capsulitis), are common conditions affecting the health of people worldwide, and have a high incidence rate among the elderly and workers engaged in repetitive shoulder tasks. In regions with scarce medical resources, achieving early and accurate diagnosis poses significant challenges, and there is an urgent need for low-cost and easily scalable auxiliary diagnostic solutions. This research introduces videos captured by consumer-grade devices as the basis for diagnosis, reducing the cost for users. We focus on the innovative application of Multimodal Large Language Models (MLLMs) in the preliminary diagnosis of shoulder disorders and propose a Hybrid Motion Video Diagnosis framework (HMVDx). This framework divides the two tasks of action understanding and disease diagnosis, which are respectively completed by two MLLMs. In addition to traditional evaluation indicators, this work proposes a novel metric called Usability Index by the logical process of medical decision-making (action recognition, movement diagnosis, and final diagnosis). This index evaluates the effectiveness of MLLMs in the medical field from the perspective of the entire medical diagnostic pathway, revealing the potential value of low-cost MLLMs in medical applications for medical practitioners. In experimental comparisons, the accuracy of HMVDx in diagnosing shoulder joint injuries has increased by 79.6\% compared with direct video diagnosis, a significant technical contribution to future research on the application of MLLMs for video understanding in the medical field.
\end{abstract}
\begin{document}

\flushbottom
\maketitle

\section{Introduction}

Shoulder disorders are becoming increasingly prevalent in modern society. The elderly and workers engaged in repetitive shoulder-based tasks, and those with prolonged sedentary desk-bound occupations constitute high-risk cohorts. According to a study by Walker-Bone K et al.~\citep{walker2004prevalence}, the prevalence of musculoskeletal pain in the upper limbs among the general population is 52\%, with shoulder pain accounting for a significant proportion. Windt DA et al.~\citep{van1996shoulder} pointed out that the annual incidence rate of shoulder diseases in general practice is approximately 14.7\%, which severely affects the quality of life of patients. Periarthritis of the shoulder, a prevalent condition characterized by pain, usually affects individuals in their fifties. In areas with limited medical resources, timely diagnosis of shoulder disorders is often lacking. Early detection is essential to prevent disease progression, reduce patient suffering, and lower treatment time and costs. Certain shoulder disorders, such as periarthritis of the shoulder, can be detected and assessed by analyzing human movement, eliminating the need for costly imaging techniques. However, research in this area remains limited.

Large Language Models (LLMs) have rapidly gained prominence in recent years, driven by a steady stream of diverse foundational models~\citep{qiu2024application,sun2024survey,firoozi2025foundation,zheng2025understanding,zheng2025sas,ren2025decoder,xu2025seqpo,ye2024reasoning,zheng2024dape}. Not only do commercial LLMs exhibit outstanding performance, but a multitude of open-source models have also achieved the state-of-the-art (SOTA) level~\citep{grattafiori2024llama}. In traditional prediction and classification tasks, supervised or unsupervised learning is the conventional approach. However, these machine learning paradigms generally require large amounts of data and rely on data scientists or algorithm engineers for model training~\citep{roh2019survey}. Given the inherent prior knowledge within LLMs, in numerous scenarios, it is not imperative to initiate LLM training from scratch. Consequently, across all industries, there is an ongoing exploration of lightweight methodologies such as prompt engineering~\citep{wei2022chain, zheng2024progressive,liu2023pre,kojima2022large,lo2024dietary,zhang2025synapseroute,zheng2023lyra}, Retrieval-Augmented Generation (RAG)~\citep{lewis2020retrieval,wei2024medco,lo2025ai}, Low-Rank Adaptation (LoRA)~\citep{hu2022lora}, and Supervised Fine-Tuning (SFT) to integrate LLMs into their respective domains, thereby fully capitalizing on the value of foundational LLMs in healthcare and medicine~\citep{qiu2023large,qiu2025emerging}.

\begin{figure}[htbp]
    \centering
    \begin{subfigure}[b]{\linewidth}
        \centering
        \includegraphics[width=\linewidth]{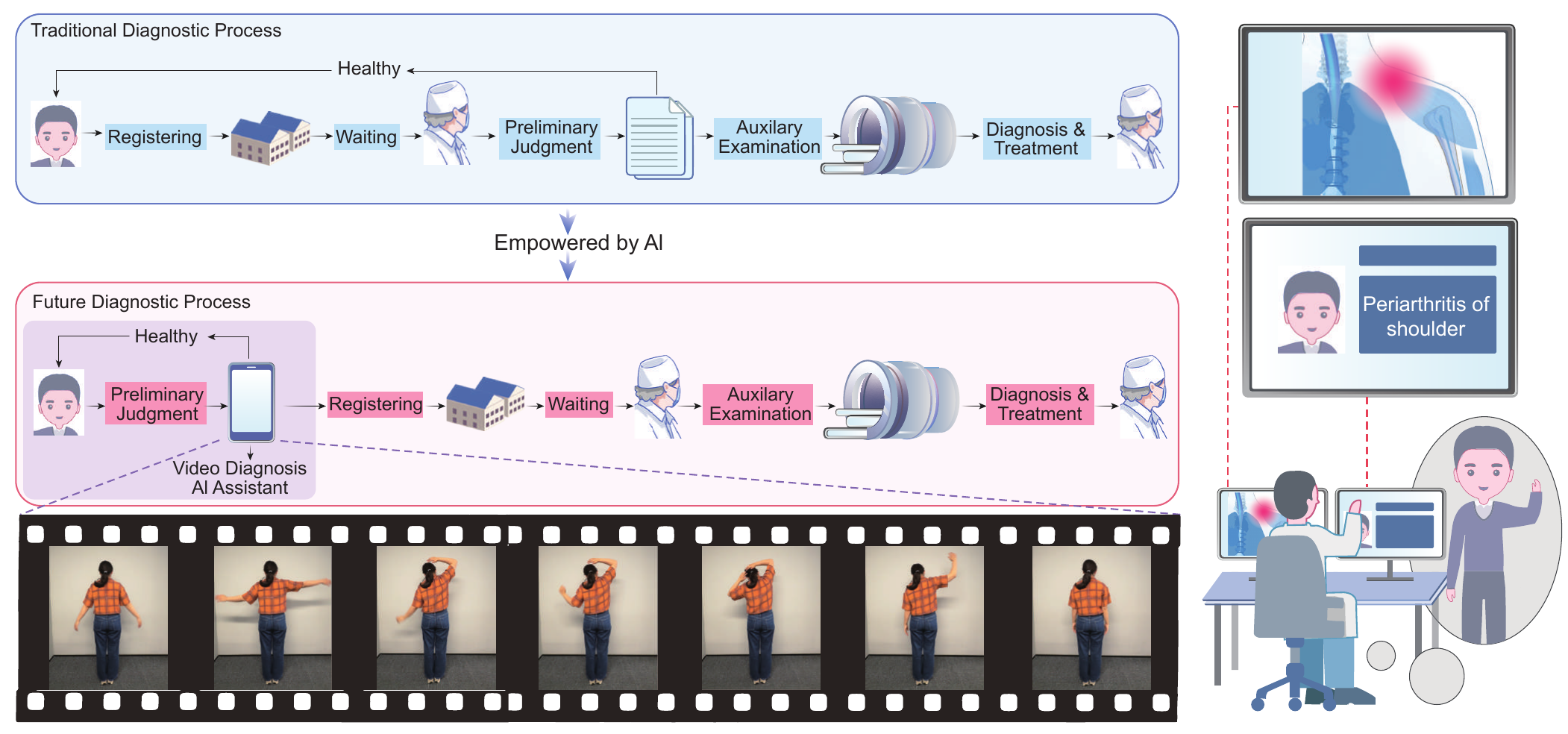}
        \caption{Diagnostic Process Comparison}
        \label{fig:diagnostic_process_comparison}
    \end{subfigure}
    \begin{subfigure}[b]{\linewidth}
        \centering
        \includegraphics[width=\linewidth]{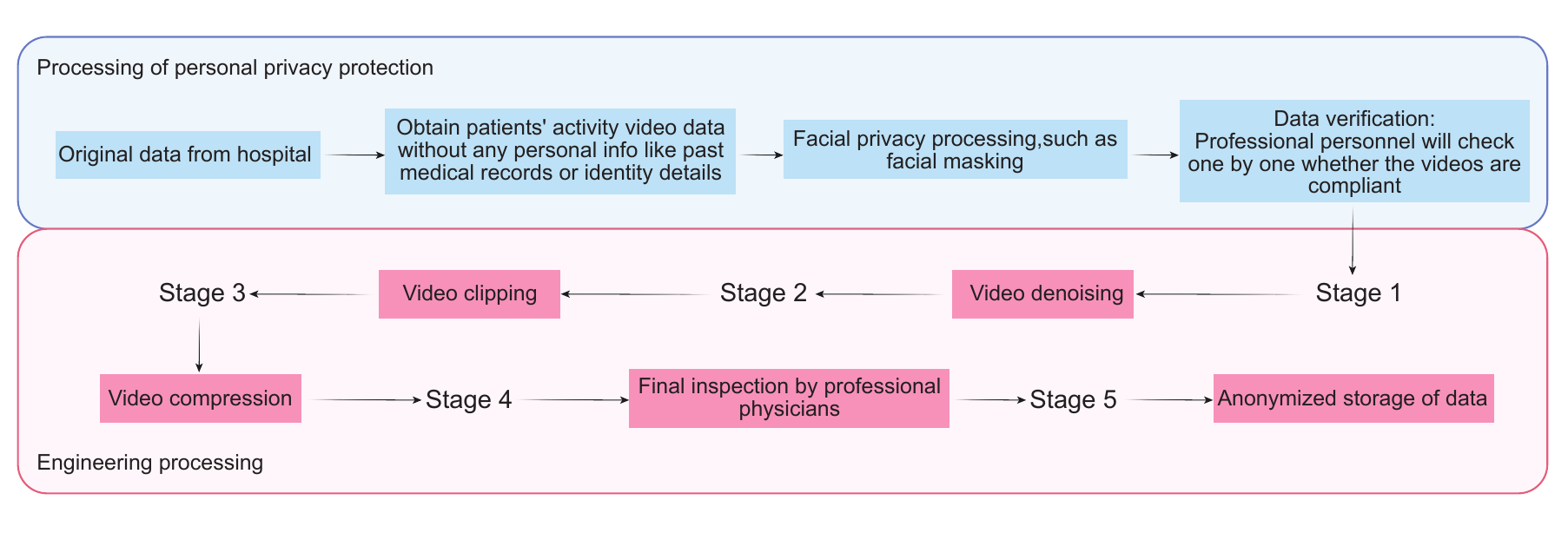}
        \caption{Data Processing Pipeline}
        \label{fig:data_processing_pipeline}
    \end{subfigure}
    \caption{\textbf{(a)} Diagnostic Process Comparison. In the traditional diagnostic process, patients must visit a hospital to determine their health status. In contrast, our envisioned AI-powered diagnostic process enables patients to assess their condition remotely using our Video Diagnosis AI Assistant, without the need to visit a hospital. This greatly reduces patient wait times and alleviates the burden on healthcare systems.
On the right, we illustrate an application scenario of our framework: a patient performs specific movements, and our model analyzes the motion video to provide a diagnosis.
\textbf{(b)} Data Processing Pipeline. Our data processing pipeline is designed with both personal privacy protection and engineering optimization in mind for handling human motion video data.}
    \label{fig:three_subfigures}
\end{figure}

This paper aims to explore a low-cost approach by leveraging prompt engineering to use MLLMs for the preliminary diagnosis of shoulder disorders. As Figure~\ref{fig:diagnostic_process_comparison} shows, we investigate whether current MLLMs, after undergoing low-cost prompt-tuning, are capable of directly conducting the preliminary diagnosis of shoulder disorders for subjects. This study also provides an affordable preliminary diagnostic solution for regions with limited medical resources. Traditionally, disease diagnosis requires patients to visit hospitals, where doctors rely on observation, inquiry, and imaging using specialized equipment to make a final diagnosis. Leveraging the visual understanding capabilities of MLLMs, we aim to facilitate early detection of shoulder disorders by analyzing the range of motion in videos recorded with consumer-grade devices. This approach streamlines the diagnostic process and improves the overall medical experience. In the future, advances in technology may enable this method to be adapted to monitor daily health in home environments, supporting long-term health tracking and timely interventions.

\section{Background and Related Works}
While recent research has demonstrated the potential of artificial intelligence (AI) in orthopedic diagnosis, spanning motion analysis to medical report interpretation, barriers remain in terms of accessibility and affordability. Zhang et al.~\citep{zhang2024combined} proposed a real-time motion evaluation system that combines MediaPipe with an improved YOLOv5 model enhanced by a Convolutional Block Attention Module, enabling more accurate detection of spinal disorders and frozen shoulder. To support rapid data processing, they adopted a client-server (C/S) architecture and developed a rehabilitation game, “Picking Bayberries,” to assist patient training. While effective, the system remains technically complex and poses significant challenges for deployment in environments without adequate AI expertise or infrastructure. In another line of work, Yu et al.~\citep{yu2023analysis} proposed an intelligent clustering algorithm based on musculoskeletal ultrasound parameters, which, combined with deep learning, enabled automatic grading of shoulder periarthritis pain. This approach provides a low-cost and non-invasive diagnostic alternative, especially for patients who cannot undergo MRI (e.g., those with metal implants), though it still relies on specialized Doppler ultrasound equipment. Vaid et al.~\citep{vaid2023using} fine-tuned a large language model (LLaMA-7B) to extract musculoskeletal pain features from unstructured clinical notes, achieving superior performance in pain localization (e.g., shoulder, lower back) and acuity classification compared to traditional NLP methods. While the study highlights the potential of LLMs in medical applications, it is primarily intended for use within medical institutions and is not readily accessible to consumers. Previous research has also explored the use of multimodal image and report data in this field. For example, Jin et al.~\citep{jin2024orthodoc} introduced the OrthoDoc model, specifically designed for the auxiliary diagnosis of Computed Tomography (CT) scans. Their results demonstrated that OrthoDoc achieved over 91\% diagnostic accuracy for common conditions, such as fractures and arthritis, outperforming both open-source and commercial models. While this study highlights the value of multimodal learning models (MLLMs) in orthopedics, it remains limited to CT data and does not incorporate dynamic motion analysis, which could be crucial for more comprehensive assessments. Similarly, Truhn et al.~\citep{truhn2023pilot} conducted a retrospective analysis of GPT-4’s potential in providing treatment recommendations for knee and shoulder conditions, using 20 anonymized MRI reports. Although their results showcase GPT-4’s effectiveness, the study is constrained by its reliance on MRI data and simplistic prompts, making it less applicable in resource-limited settings where such diagnostic tools may not be readily available.

In the accurate diagnostic process of diseases, imaging report and data such as those from ultrasound, CT, or MRI remain indispensable and crucial evidence. However, considering that these professional examinations are often accompanied by high costs and resource limitations, this study aims to explore an innovative method that is cost-effective and widely accessible. 
Our goal is to harness video data from consumer-grade devices, such as smartphones and home cameras, to enable the diagnosis of shoulder disorders using MLLMs. This not only has the potential to lower the cost of medical services but can also significantly enhance the accessibility of preliminary disease screening services.

In this research, we explore the feasibility of the direct application of MLLM in the field of the preliminary diagnosis of shoulder disorders. To our knowledge, this is the first work using videos captured by low-cost pervasive consumer-level cameras and MLLMs to diagnose shoulder disorders. Our contributions are threefold:
\begin{itemize}
    \item In the diagnostic framework, aiming at the problem of information loss in the direct judgment made by MLLMs, we have innovatively proposed Hybrid Motion Video Diagnosis (HMVDx) and found an extremely low-cost implementation method. This method allows the MLLM to be responsible for converting video information into action description texts, and a reasoning large language model makes a judgment based on these descriptions and pre-set diagnostic rules. This division of labor reduces the complexity of the model's tasks and improves the accuracy and reliability of diagnosis. Experiments show that this method has clear advantages when dealing with the data of patients with shoulder disorders.
    \item For diseases that can be preliminarily diagnosed by observing the range of body movements, we propose a Motion Trajectories Prompt Framework. This enables LLMs to achieve understanding of human actions, which helps in making more accurate judgments subsequently. This framework utilizes Gemini-1.5-Pro to analyze orthopedic popular science videos, summarize the judgment actions and standards, and replace numerical quantification descriptions with relative position descriptions to improve accuracy. In the future, the framework can help medical practitioners transform general LLMs into specific ones at a low cost.
    \item In view of the limitations of traditional evaluation indicators when MLLMs in the field of orthopedics are used for the preliminary diagnosis of shoulder disorders, we constructed an innovative evaluation system. It integrates traditional metrics with Usability Index, a novel metric proposed in this work that accurately evaluates the model from three dimensions: the integrity of action recognition, the rationality of behavior judgment, and the accuracy of the final judgment. It cooperates with detailed scoring standards to disassemble and analyze the model's output, and sets up a three-level filtering scenario to strengthen the constraints, and deeply analyzes the model's capabilities. This system provides a scientific and comprehensive tool for analyzing the applicability of MLLMs in the preliminary diagnosis of shoulder disorders, complementing the traditional metrics. 
\end{itemize}

\begin{figure}[htbp]
    \centering
    \begin{subfigure}[b]{0.3\textwidth}
        \centering
        \includegraphics[width=\linewidth]{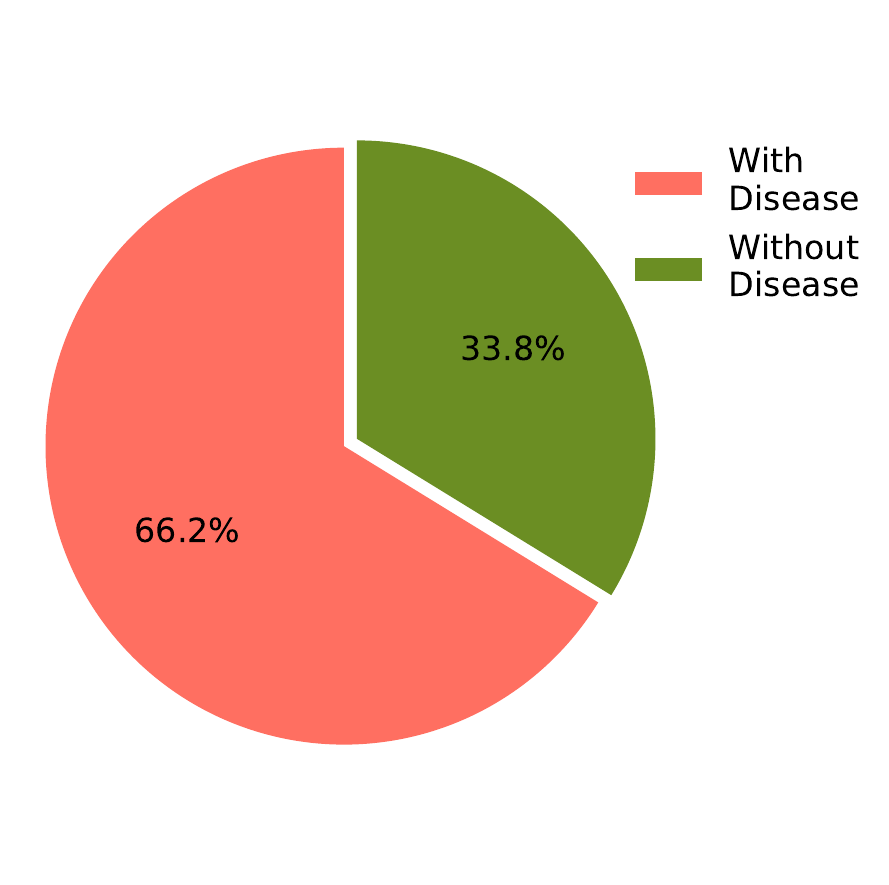}
        \caption{Disease Distribution}
        \label{fig:sub1}
    \end{subfigure}
    \hfill
    \begin{subfigure}[b]{0.3\textwidth}
        \centering
        \includegraphics[width=\linewidth]{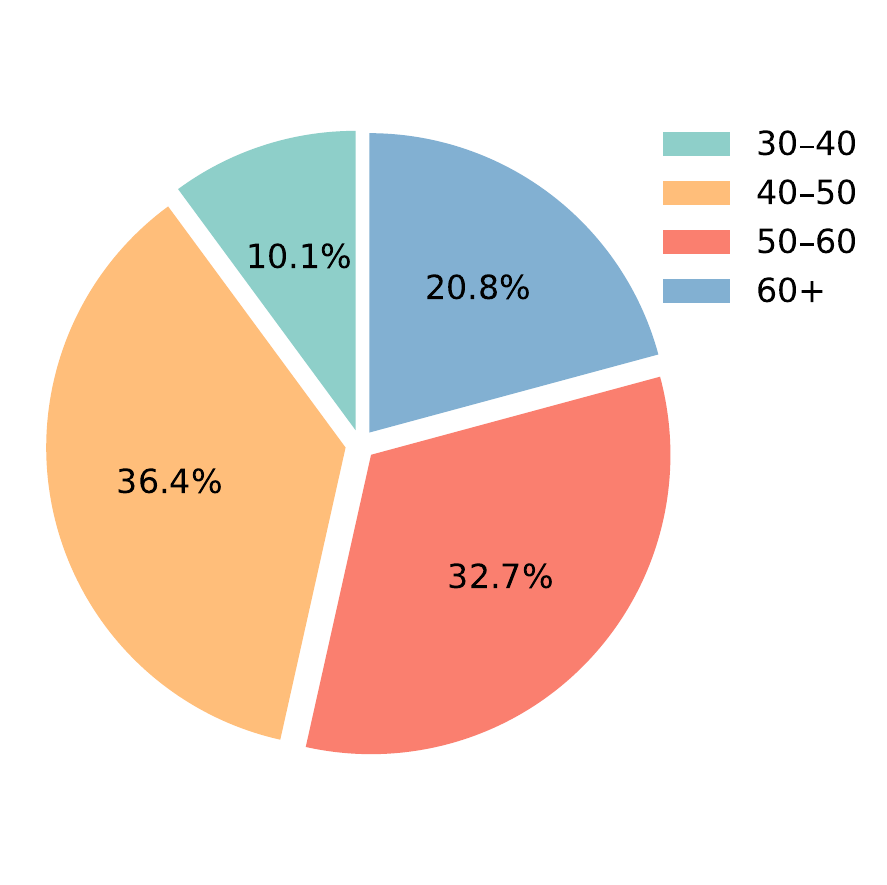}
        \caption{Age Distribution}
        \label{fig:sub2}
    \end{subfigure}
    \hfill
    \begin{subfigure}[b]{0.3\textwidth}
        \centering
        \includegraphics[width=\linewidth]{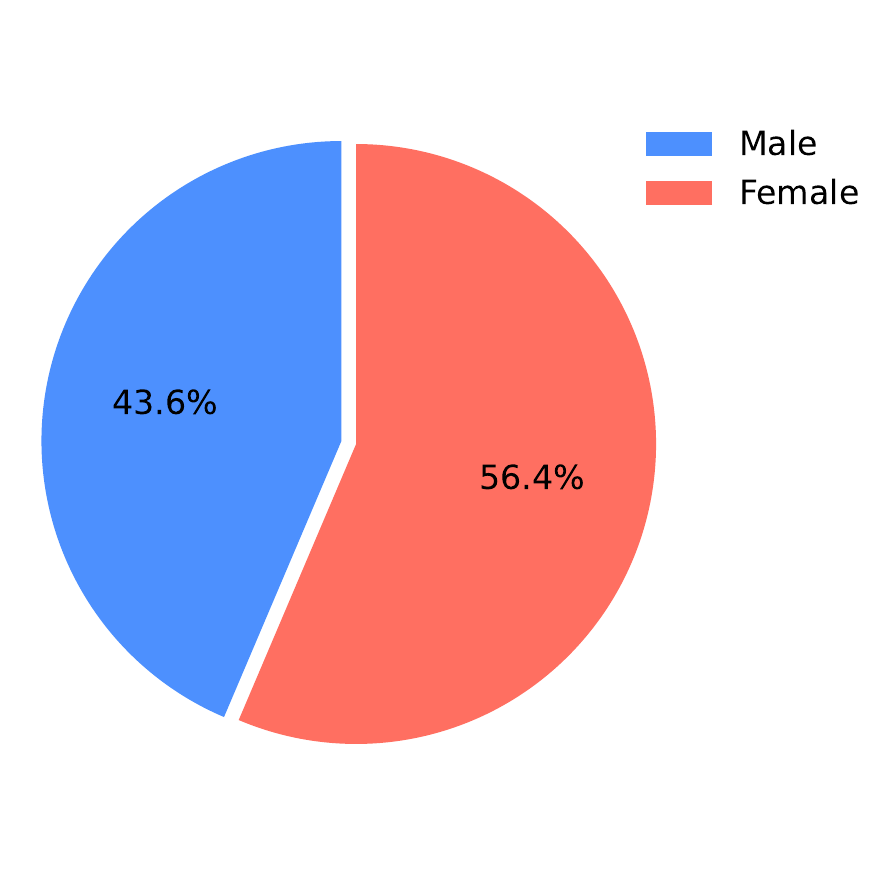}
        \caption{Gender Distribution}
        \label{fig:sub3}
    \end{subfigure}
    \caption{Pie charts illustrating the distribution of disease status, age groups, and gender in our dataset.}
    \label{fig:dataset_distribution}
\end{figure}

\section{Results}

\begin{figure}
    \centering
    \includegraphics[width=\linewidth]{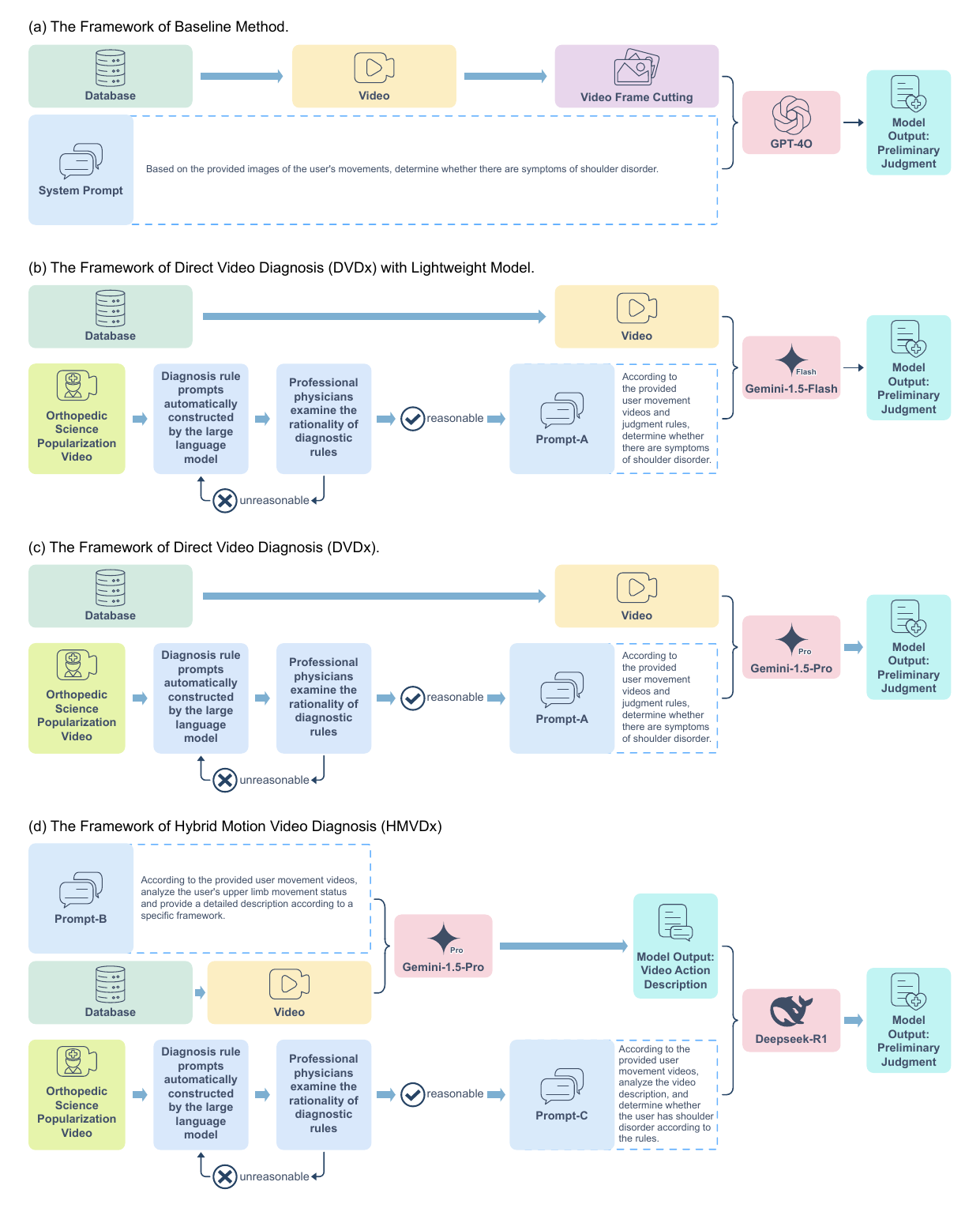}
    \caption{Overview of Framework Variants for Video Diagnosis}
    \label{fig:video_diagnosis_frameworks}
\end{figure}

In this study, we use videos captured by consumer-grade devices as the input of MLLMs, and apply them to the preliminary determination and identification of shoulder disorders. 
Therefore, to accurately quantify the performance gain brought by the additional temporal information when upgrading from image to video modality, we selected GPT-4o~\citep{hurst2024gpt} as a baseline. 
The process is shown in Figure~\ref{fig:video_diagnosis_frameworks}(a). We process the same batch of samples into a certain number of image sequences, which are used as the input of the GPT-4o model for disease diagnosis. Although GPT-4o has strong multimodal understanding capabilities across text, audio, and images, it does not natively support video as a temporally coherent modality. While it is technically possible to input videos to GPT-4o by feeding frames sequentially, this frame-by-frame processing lacks explicit temporal modeling and continuity understanding, which are critical for capturing dynamic movement patterns in diagnostic tasks. In our evaluations, GPT-4o's performance on such frame-wise video approximations was lower than that of models designed with dedicated video understanding capabilities. This supports our decision to treat GPT-4o as a control to isolate the impact of temporal information. 
Meanwhile, in the application of MLLMs, we also conduct a comparative study on the diagnostic effects of Gemini-1.5-Flash and Gemini-1.5-Pro (See Figure~\ref{fig:video_diagnosis_frameworks}). The aim is to explore whether there are significant differences in the diagnostic effects on shoulder disorders between the lightweight model and more complex models. By transferring the core knowledge of Gemini-1.5-Pro to a smaller model, Gemini-1.5-Flash achieves a lightweight architecture while maintaining its multimodal capabilities. Its reasoning speed has been significantly improved, and it is also capable of handling the understanding of long videos.
Therefore, in result evaluation, we will divide it into two modules. Firstly, we conduct a comparative analysis of the effectiveness of the GPT-4o model based on image input, Direct Video Diagnosis (DVDx), and the HMVDx proposed in this study. The aim is to analyze the incremental information brought by the video modality and the optimization space of HMVDx. Secondly, in DVDx, we compare the performance differences of models with different sizes to determine whether there are significant differences in the cost of the models and the complexity of their architectures for the diagnostic task.

\subsection{Quantitative Analysis}
\subsubsection{Analysis of Comprehensive metrics}
\begin{figure} 
    \centering
    \begin{subfigure}[b]{0.48\linewidth}
        \centering
        \includegraphics[width=\linewidth]{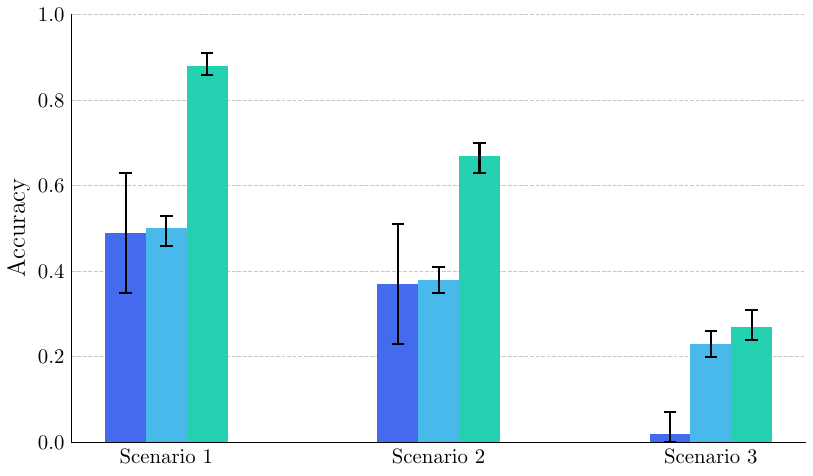}
        \caption{Comparison of Methods - Accuracy}
        \label{fig:subfig_methods_accuracy}
    \end{subfigure}
    \hfill
    \begin{subfigure}[b]{0.48\linewidth}
        \centering
        \includegraphics[width=\linewidth]{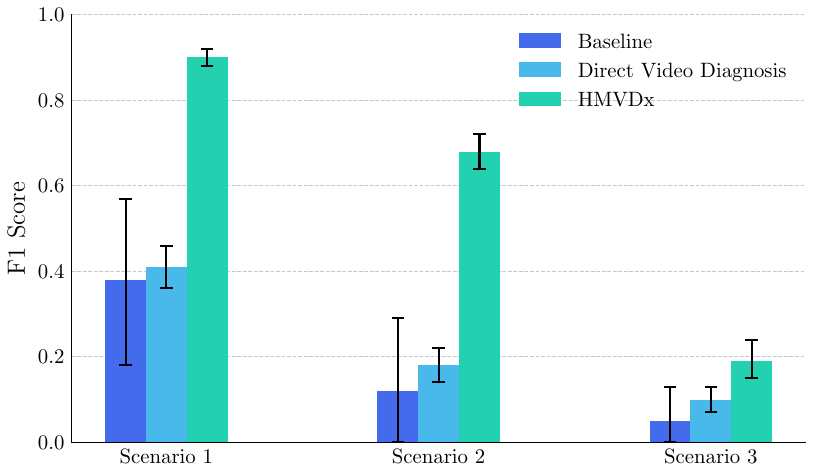}
        \caption{Comparison of Methods - F1 Score}
        \label{fig:subfig_methods_f1}
    \end{subfigure}
    \vskip\baselineskip
    \begin{subfigure}[b]{0.48\linewidth}
        \centering
        \includegraphics[width=\linewidth]{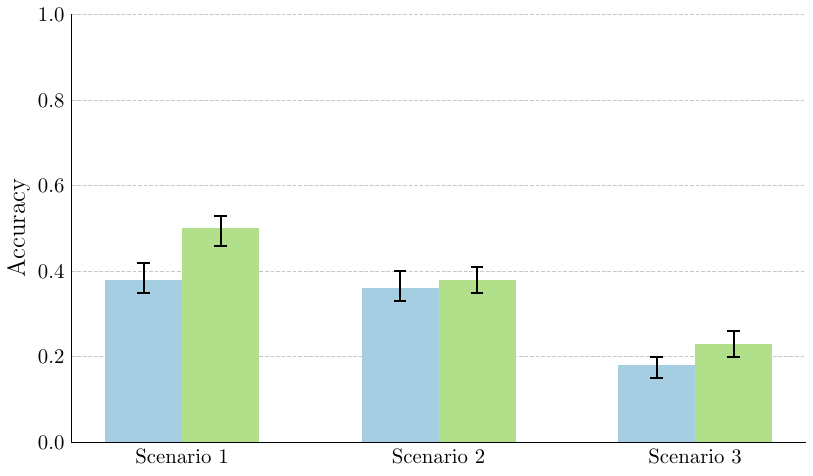}
        \caption{Comparison of Model Size - Accuracy}
        \label{fig:subfig_size_accuracy}
    \end{subfigure}
    \hfill
    \begin{subfigure}[b]{0.48\linewidth}
        \centering
        \includegraphics[width=\linewidth]{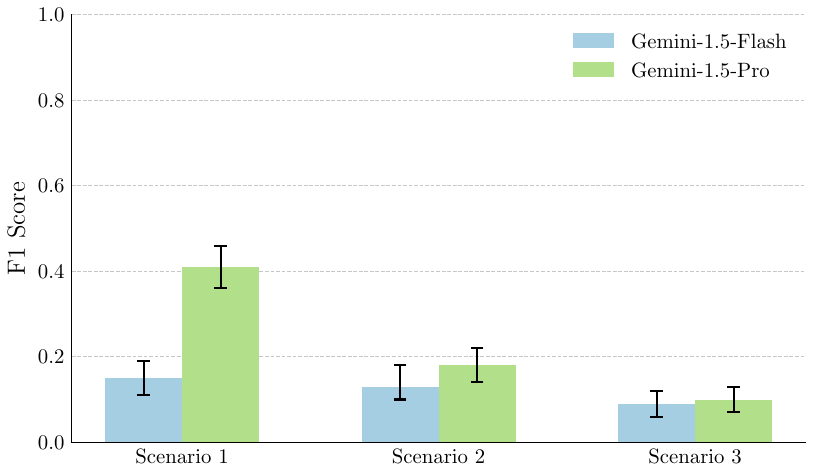}
        \caption{Comparison of Model Size - F1 Score}
        \label{fig:subfig_size_f1}
    \end{subfigure}
    \vskip\baselineskip
    \begin{subfigure}[b]{0.48\linewidth}
        \centering
        \includegraphics[width=\linewidth]{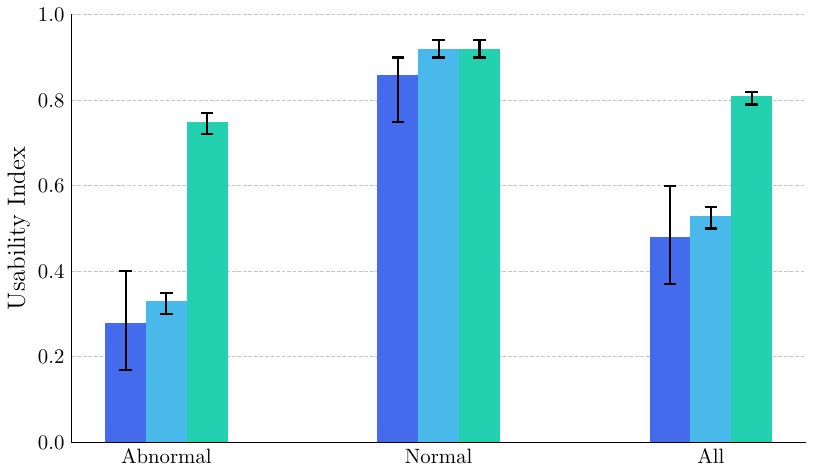}
        \caption{Comparison of Methods - UI}
        \label{fig:comparision_methods_ui}
    \end{subfigure}
    \hfill
    \begin{subfigure}[b]{0.48\linewidth}
        \centering
        \includegraphics[width=\linewidth]{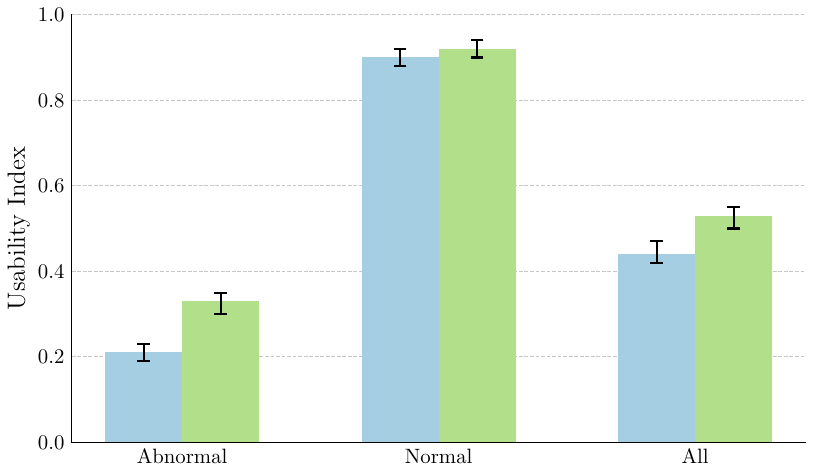}
        \caption{Comparison of Model Size - UI}
        \label{fig:comparision_size_ui}
    \end{subfigure}
    \caption{Comparison of different methods and model sizes in terms of Accuracy, F1 Score, and UI.}
    \label{fig:all_comparisons}
\end{figure}

Under different constraint scenarios, we conducted a systematic analysis of comprehensive evaluation metrics. These metrics cover multiple dimensions, such as Accuracy, Precision, Recall, and F1 score. We aim to conduct a comprehensive evaluation of the performance of three approaches, including Baseline, Direct Video Diagnosis, and Hybrid Motion Video Diagnosis (Figure~\ref{fig:video_diagnosis_frameworks}), in diagnosing shoulder disorders. Table~\ref{tab:comprehensive_matrix} presents a full summary of the system's performance across all evaluation metrics. We designed three scenarios to evaluate the methods, and their performance across these scenarios is shown in Figure~\ref{fig:subfig_methods_accuracy} and Figure~\ref{fig:subfig_methods_f1}.

In Scenario 1, that is, the bottom-line constraint scenario where only the final judgment result is concerned, HMVDx demonstrated excellent performance in this scenario. Its accuracy rate reached 0.88, and the F1 score was 0.90, with the recall rate and precision rate being 0.83 and 0.98, respectively. Specifically, the accuracy rate of HMVDx was 79.6\% and 76.0\% higher than that of the GPT-4o baseline and Direct Video Diagnosis (DVDx) methods respectively, and the F1 score was 136.8\% and 119.5\% higher respectively.  
Overall, HMVDx significantly outperforms both GPT-4o and DVDx in terms of diagnostic performance, demonstrating its ability to deliver highly accurate final judgments. From the perspective of medical diagnosis, a high accuracy rate means that this method can correctly diagnose a high proportion of cases in a large number of samples, and a high F1 score comprehensively reflects a balance between precision and recall, indicating that HMVDx performs excellently both in identifying positive cases and avoiding misdiagnoses. In contrast, the GPT-4o baseline and DVDx performed poorly in this scenario.

Scenario 2 is the logical constraint scenario that takes into account both the final judgment and the rationality of the behavior judgment, HMVDx still significantly outperformed the GPT-4o baseline and DVDx in various metrics. Especially in terms of the F1 score, the GPT-4o baseline and DVDx had a substantial decline, indicating that these two models are infeasible under the constraint of logical consistency. The F1 score of HMVDx was approximately 4.7 and 2.8 times higher than that of the GPT-4o baseline and DVDx, reaching 0.68. Generally, an F1 score of around 0.7 is considered a good result.

Scenario 3, as the most stringent whole-process constraint scenario, comprehensively considers the integrity of action recognition, the rationality of behavior judgment, and the final judgment. In this scenario, although HMVDx was still superior to the GPT-4o baseline and DVDx, the comprehensive indicators were not satisfactory. Taking the F1 score as an example, that of HMVDx was only 0.19. The low F1 score of HMVDx in Scenario 3 indicates that under the strict requirements of the whole process, the model also has deficiencies. It is worth noting that after analyzing the reasons for the bad cases of HMVDx, it was found that the main reason was that the MLLMs for visual understanding in the first step had biases or omissions in the description of human actions. In Scenario 3, the performance of the GPT-4o baseline was significantly lower than that of DVDx, indicating that for the task of identifying and judging human actions by MLLMs, the modal change from images to videos still brings valuable incremental information. 

As Figure~\ref{fig:subfig_size_accuracy} and Figure~\ref{fig:subfig_size_f1} shows, by comparing the performance of Direct Video Diagnosis based on models of different sizes, we found that both the F1 score and accuracy of Gemini-1.5-Flash are lower than those of Gemini-1.5-Pro. Especially in Scenario 1, the accuracy and F1 score of Gemini-1.5-Pro are 31.58\% and 173.33\% higher than those of the lightweight model. Therefore, in the comparison of the diagnostic effects of models on shoulder disorders, we found that more complex models with larger parameters have advantages.

Considering the performance of different scenarios and models comprehensively, HMVDx outperforms the GPT-4o baseline and DVDx in all scenarios, demonstrating certain advantages. However, the performance of HMVDx drops in Scenario 3, which means that this method cannot provide reliable judgment results when facing the strict constraint of the whole process. Scenario 3 is the closest to a realistic medical scenario, covering a series of key processes such as recognition, judgment, and summarization. If Scenario 3 is taken as the standard, the current three methods cannot meet the requirements of the real medical scenario and are difficult to implement in the actual medical environment. This is because in actual medical scenarios, the reliability of the diagnostic results is of crucial importance, and any mistake in any stage may have a serious impact on the patient's health.

Nevertheless, we still emphasize the huge potential of HMVDx. In Scenario 2, on the premise that all judgments are reasonable, the F1 score of HMVDx still reaches 0.68. Considering the difficulty of judging shoulder disorders, this is a fairly decent level, indicating that HMVDx can achieve a good balance between precision and recall when meeting certain requirements of logical rationality, and has potential to be applied in the field of preliminary judgment of shoulder disorders. The reason is that HMVDx divides recognition and diagnosis into two tasks, fully leveraging individual advantages of the MLLM and the reasoning model. Gemini-1.5-Pro focuses on the description of human actions, while the DeepSeek-R1 can more effectively make reasonable judgments on action descriptions, thus achieving better results. It is worth noting that none of the three methods employed advanced supervised fine-tuning (SFT) or other fine-tuning techniques; instead, they relied solely on the inherent capabilities of the MLLMs. This means that if more advanced optimization technologies are introduced in the future, HMVDx is expected to further improve its performance in the whole-process constraint scenario, thus getting closer to the requirements of actual medical applications. 

\begin{table}[htbp]
    \centering
    \caption{Usability index}
    \resizebox{\textwidth}{!}{
    \begin{tabular}{cccccc}
    \toprule
    \textbf{Framework} & \textbf{Dimension} & \textbf{Mean} & \textbf{95\% CI - Lower Bound} & \textbf{95\% CI - Upper Bound} \\
    \midrule
\multirow{3}{*}{\parbox{3.5cm}{\centering Baseline}} &	Normal &	0.856&	0.750&	0.900\\
	&Abnormal&	0.282	&0.170&	0.402\\
	&Overall	&0.481	&0.372	&0.597\\
    \hline
\multirow{3}{*}{\parbox{3.5cm}{\centering Direct Video Diagnosis with Gemini-1.5-Flash}} &	Normal&	0.899	&0.877&	0.919\\
	&Abnormal&	0.209&	0.189&	0.233\\
	&Overall&	0.443&	0.415 & 0.472 \\
    \hline
\multirow{3}{*}{\parbox{3.5cm}{\centering Direct Video Diagnosis with Gemini-1.5-Pro}} &	Normal&	0.922	&0.904	&0.941 \\
	&Abnormal&	0.326&	0.299&	0.354\\
	&Overall&	0.528	&0.502&	0.554\\
    \hline
\multirow{3}{*}{\parbox{3.5cm}{\centering Hybrid Motion Video Diagnosis}}	&Normal	&0.922&	0.901&	0.938 \\
	&Abnormal&	0.747&	0.722&	0.770\\
	&Overall&	0.806&	0.787&	0.823\\
    \bottomrule
    \end{tabular}
    }
    \label{tab:supp_usability_index}
\end{table}

\subsubsection{Analysis of Usability Index}

\begin{table}[htbp]
    \centering
    \caption{Summary of grading rubric for model output}
    \resizebox{\textwidth}{!}{
    \begin{tabular}{m{4cm}m{4cm}m{7cm}}
        \toprule
        \textbf{Label Classification} & \textbf{Label Rules} & \textbf{Label Definition} \\
        \midrule
        \multirow{3}{*}[-3ex]{\parbox{4cm}{\centering Integrity of Movement Recognition (A, three-class classification from 0 to 1)}} & 

1 = Complete Recognition & 
This method can accurately capture all user actions with clinical indicative significance in the video, such as abduction, forward flexion, internal rotation and other standard actions, without any omission. \\
\cline{2-3}
& 0.5 = Partial Recognition & 
This method can only capture some of the key actions. \\
\cline{2-3}
& 0 = Severe Lack & 
This method completely omits all core actions and fails to effectively recognize the key actions. \\

\hline
\multirow{3}{*}[-10ex]{\parbox{4cm}{\centering Rationality of Movement Judgment (R, three-class classification from 0 to 1)}}
&

1 = Completely Rational &
The accurate and reasonable judgments made by this method for each action are highly consistent with the pre-set judgment rules, with rigorous and reasonable logic. \\
\cline{2-3}
& 0.5 = Partially Rational & 
The judgments of some actions by this method conform to the rules, but in the overall logical deduction process, there may be some non-critical logical leaps or misjudgments. \\
\cline{2-3}
& 0 = Completely Irrational & 
The judgment of the core actions by this method is contrary to the content of the original video. For example, in the original video, the patient obviously cannot touch the back with the hand behind, but this method mistakenly describes that the patient can flexibly perform the action of touching the back with the hand behind. \\
\hline
\multirow{2}{*}[-5ex]{\parbox{4cm}{\centering Accuracy of Final Judgment\\(D, binary classification)}}
 & 
1 = Correct Diagnosis & 
The final diagnosis result given by this method is completely consistent with the result after being reviewed by professional doctors in the hospital. \\
\cline{2-3}
& 0 = Incorrect Diagnosis & 
It covers the situations of missed diagnosis (wrongly judging an actually positive case as negative), misdiagnosis (wrongly judging an actually negative case as positive), and the situation where this method cannot give an effective result. \\
        \bottomrule
    \end{tabular}
    }
    \label{tab:grading_rubric}
\end{table}

Due to the definition of the Usability Index (UI), its values range from 0 to 1, with higher values indicating better model usability. As shown in Figure~\ref{fig:comparision_methods_ui},~\ref{fig:comparision_size_ui} and Table~\ref{tab:supp_usability_index}, the overall performance of the GPT-4o model is relatively poor, achieving a UI of only 0.48. This underperformance can be attributed to GPT-4o’s exclusive reliance on image inputs, which severely limits its ability to understand motion and interpret actions effectively. In contrast, the HMVDx model demonstrates strong performance, with a UI of 0.81, substantially higher than that of DVDx (0.53), indicating a clear advantage in overall usability. Further analysis by sample type reveals that for positive cases (i.e., patients with shoulder joint disorders), HMVDx achieves a UI of 0.75, far outperforming the GPT-4o baseline (0.28) and DVDx (0.33). This suggests that HMVDx is significantly more effective at recognizing patient movements, interpreting clinical behavior, and producing accurate diagnostic outcomes. For negative cases (i.e., healthy users), HMVDx attains a UI of 0.92, which is comparable to DVDx and slightly higher than GPT-4o’s 0.86. This indicates that for non-injured individuals, the performance gap among the three methods is relatively minor.

By comparing the performance of DVDx based on models of different sizes, we found that the UI of Gemini-1.5-Pro is higher than that of Gemini-1.5-Flash (0.53 versus 0.44), indicating that more complex models still have certain advantages in the tasks of action recognition and judgment.

Overall, HMVDx has demonstrated significantly better usability performance than the GPT-4o baseline and DVDx when recognizing shoulder joint injuries. There may be various reasons behind this result. Diagnosing shoulder joint injuries usually involves several key steps, including action recognition, behavior diagnosis, and final judgment. Both the GPT-4o baseline and DVDx use a single MLLM for judgment. This means that the model needs to simultaneously undertake multiple tasks, such as accurately recognizing actions from complex videos, making reasonable behavior diagnoses based on the actions, and providing accurate final judgment results. This task allocation method places extremely high demands on the comprehensive capabilities of the model and increases the likelihood of errors or biases in the model. In contrast, HMVDx creatively employs two models to work in a division-of-labor and collaborative manner. Gemini-1.5-Pro focuses on providing detailed descriptions of the actions of people in the video, transforming complex video information into an understandable text format. The DeepSeek-R1 model, on the other hand, makes judgments based on these text descriptions. This division-of-labor model effectively reduces the task complexity of each model, enabling each model to exert its maximum effectiveness, thus improving the accuracy and reliability of the entire diagnostic process. 

\subsection{Qualitative Analysis}
\begin{figure}
    \centering
    \includegraphics[width=\linewidth]{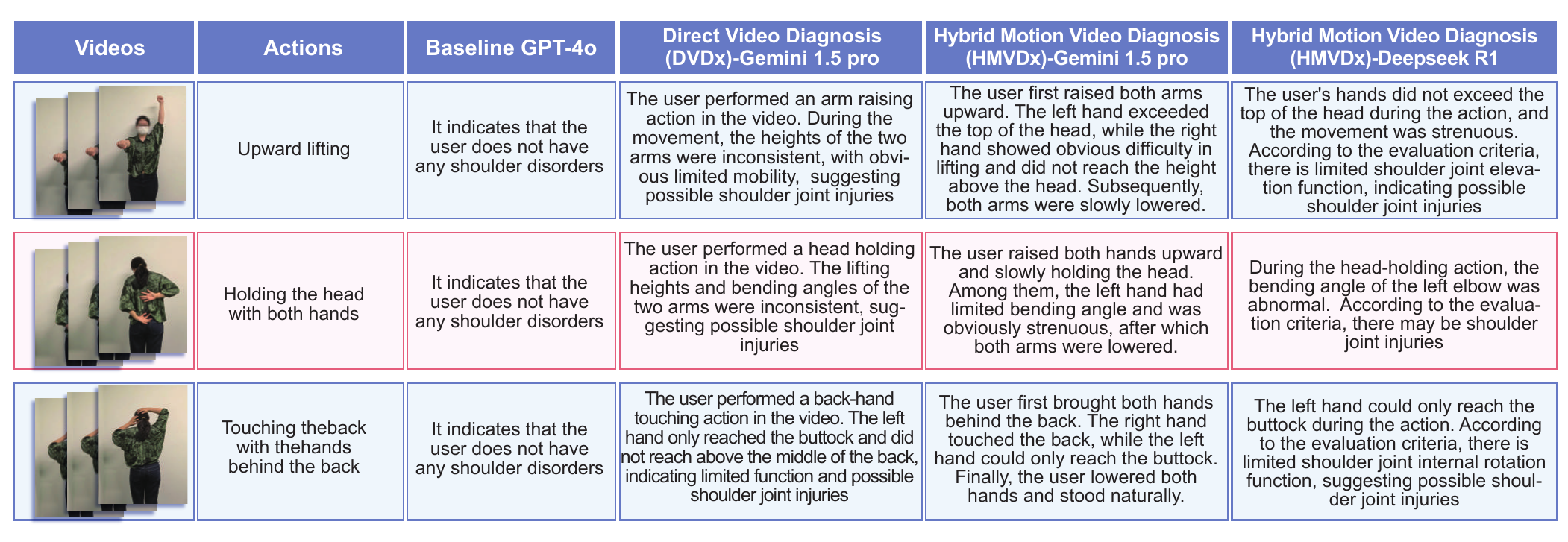}
    \caption{Qualitative Results. Our method demonstrates superior performance over GPT-4o and DVDx}
    \label{fig:shoulder_disorder_annotation_description}
\end{figure}

In the analysis of the experimental results, we found some interesting aspects in  MLLM's understanding of human actions.
There may be certain deviations in the description and understanding of free actions. 
Since we did not constrain the action descriptions generated by the MLLM, the model often produced non-standardized, free-form descriptions that included compound actions (e.g., circular motions, arm swings). However, As Figure~\ref{fig:shoulder_disorder_annotation_description} shows, these compound actions are difficult for the reasoning model to decompose into standard evaluative elements (e.g., upward lifting, backward extension), making it challenging to establish effective diagnostic logic.

There is also a problem of deviation in judging the starting and ending points during action decomposition. 
For example, experimental analysis shows that when a target action involves multiple sub-steps (e.g., "putting hands on the head" requires first raising the arms and then placing the hands behind the head), the model often struggles to distinguish between the core action and its individual components.
This phenomenon is due to unawareness of the action intention during the video understanding process, leading to the model misjudging the process action as the formal action in the clinical action evaluation scenario (such as the arm backward extension test). 

The recognition of spatial positioning is of crucial significance for judgment. In the model output, we found two types of problems: First, it is challenging to accurately assess limb verticality from a front-facing view. For example, it is difficult to determine whether the arm is fully perpendicular to the ground. In particular, the recognition of the external rotation action and the judgment of its range involve the understanding of spatial depth, causing the performance of MLLMs in this regard is not satisfactory; Second, the recognition rate of the human trunk part is relatively low from the back view (especially the misrecognition rate of the waist area is relatively high). In the optimization of the prompt, for the recognition of the back position, we have improved the recognition accuracy to a certain extent by emphasizing prompts such as paying attention to the waist and hips. 
Finally, the visual model has a relatively high error rate in judging the left and right directions of the human body.

Overall, qualitative results show that HMVDx provides more accurate and interpretable assessments of shoulder joint injuries compared to GPT-4o and DVDx. It better captures complex actions, spatial relationships, and subtle posture variations, making it more reliable for real-world diagnostic use.

\section{Discussion}

In summary, this study explores the application of MLLMs to the preliminary diagnosis of shoulder disorders through the Motion Trajectories Prompt Framework. An innovative HVMDx method has been proposed, and a corresponding evaluation system has been constructed. Through quantitative and qualitative analysis of different methods, we have found that HMVDx demonstrates significant advantages in the preliminary diagnosis of shoulder disorders. By means of division of labor and cooperation, it reduces the task complexity of a single model and improves the accuracy and reliability of judgment. However, although HMVDx performs outstandingly in terms of performance, all current methods still cannot meet the strict requirements of practical medical applications according to Scenario 3, which is the closest to the real medical scenario. More statistical comparison on research is available in Supplementary Materials.

Currently, MLLMs have problems in aspects such as dynamic video analysis, body positioning accuracy, action decomposition accuracy, and orientation recognition, which indicate the direction of optimization for future research. This study has revealed the potential of low-cost MLLMs in medical applications for assisting medical practitioners with diagnosing shoulder disorders. Moreover, it has made contributions in the construction of medical motion prompt framework, diagnostic methods, and improvement of the evaluation metrics when MLLMs are applied to the preliminary diagnosis of shoulder disorders.

The core research objective is to verify whether MLLMs have the capability to construct practical preliminary diagnostic tools through low-threshold prompt tuning. Demonstrating the technical feasibility of this approach is crucial to assessing its potential for widespread adoption in primary care settings. Experimental comparisons have found that a series of diseases that can be interpreted based on visual representations (such as the assessment of muscle group status in lumbar muscle strain and the identification of local swelling characteristics in tenosynovitis) will have the opportunity to be transformed into standardized intelligent screening tools. Due to its low dependence on professional annotation data and computing power, this technical path particularly conforms to the core concern of this study regarding the technical scalability. It is expected to alleviate the uneven distribution of primary medical resources by constructing a lightweight diagnostic system and providing scalable technical support for improving the level of group health management. Future research can expand and deepen from multiple dimensions on the basis of this study. Technologies such as Supervised Fine-Tuning (SFT), Retrieval Augmented Generation (RAG), Agentic AI~\citep{qiu2024llm}, and knowledge graphs should be actively introduced to further explore the application potential of MLLMs in the medical field. Furthermore, research could be carried out in a multilingual environment to enable the model to adapt to medical data with different languages and broaden the application scope of the research findings. At the data level, more high-quality and diverse medical imaging and video data could be leveraged, covering various demographics, further expanding the scale of the dataset and improving the generalization ability of the model and the accuracy of preliminary diagnosis.

\section{Methods}

\subsection{Real-world Datasets for \algoName}

This research was conducted in close collaboration with hospitals and medical institutions, which contributed a rich and targeted dataset of human motion videos for this study. These videos cover the movement of individuals with shoulder disorders, as well as activity videos of healthy people.

We applied personal privacy protection and engineering processing to the human motion video data using the pipeline illustrated in Figure~\ref{fig:data_processing_pipeline}. In order to comply with the principles of privacy protection, after obtaining sample videos of human movement, we have taken comprehensive privacy protection measures. For the personal information involved in the videos, including identification information and sensitive privacy content such as medical histories, all have been concealed and eliminated through professional technical means. At the same time, to further safeguard the facial privacy security of the people in the videos, we have employed techniques such as facial masking and blurring to ensure that individual facial features cannot be recognized.
During data pre-processing, in order to enable the MLLMs for video understanding to focus more accurately on the understanding of human movements, we eliminated the sound from the video.
Meanwhile, we select videos with independent and complete human movements to reduce the impact of irrelevant information on the model's evaluation. To increase the efficiency of model inference, we cropped videos and compressed their size to make them more flexible.

To ensure high-quality annotations, we invited some distinguished medical experts specializing in orthopedics to manually annotate the samples.When assessing the movement around the shoulder joint, the evaluation mainly focuses on aspects such as whether there are limitations in actions like the upward lifting, external rotation and abduction, internal rotation and adduction of the affected upper limb, and whether the range of motion has decreased. Meanwhile, in cases where there is a comparison of bilateral movements, the emphasis is placed on comparing whether the movement performance of the affected side and the healthy side is balanced or shows differences. For example, the upward lifting range of the affected upper limb is significantly smaller compared to that of the healthy side. In addition, the experts will also make annotations for situations in the video samples where it is difficult to make an accurate judgment due to incomplete movements. Here we provide a description of the sample. For instance, a person performed three actions in total: upward lifting, holding the head with both hands, and touching the back with the hands behind the back, which involve functions such as upward lifting, internal rotation, external rotation, adduction, and abduction of the shoulder joint.  The upward lifting range of the affected side is obviously smaller than that of the healthy side. The healthy upper limb can be lifted up to 180° (perpendicular to the ground), while the upward lifting range of the affected side is approximately between 90° and 120° (slightly higher than the horizontal position). When the person performs the action of holding the head with the affected upper limb, the height is relatively low, and when performing the action of touching the back with the hands behind the back, the affected upper limb can only be lifted to the height of the buttocks, indicating that the functions of abduction and external rotation (holding the head with both hands) and internal rotation and adduction (touching the back with the hands behind the back) are limited.
It can be inferred that the patient has a unilateral shoulder joint disorder.
Based on the above criteria, we have sought to obtain the maximum number of samples possible. 
The final sample dataset consists of a total of 761 sample videos. After being labeled by doctors, 504 samples are highly likely to indicate shoulder disorders, while 257 samples are highly likely to have no shoulder disorders. The distributions of disease status, age groups, and gender within our dataset are presented in Figure~\ref{fig:dataset_distribution}.

\subsection{HMVDx}
In a real diagnostic environment, doctors establish a basic understanding of the patient's movement by observing the patient's limb activities, and then make a judgment based on medical knowledge as to whether there may be abnormal diseases. Therefore, we simulate this process and propose a Direct Video Diagnosis (DVDx) method and Hybrid Motion Video Diagnosis (HMVDx) based on motion understanding and disease diagnosis.

Direct Video Diagnosis (DVDx) involves inputting video samples into a MLLM and then directly obtaining the diagnostic results. Specifically, the operation is as follows: as Figure~\ref{fig:video_diagnosis_frameworks} shows, first, the system prompt is finely tuned, and then the video samples are input into the MLLM, expecting the model to output diagnostic results based on the understanding of the video. We employ Gemini-1.5-Pro as the core model for direct video diagnosis, as it supports long-context understanding and exhibits strong reasoning capabilities in complex video tasks~\citep{team2024gemini}. It can effectively integrate video, audio, and text features. In the EgoSchema task~\citep{mangalam2023egoschema}, Gemini-1.5-Pro achieved an accuracy rate of 70.2\% using only 16 frames, setting a new SOTA performance (in comparison, the accuracy rate of GPT-4V is 55.6\%). Meanwhile, in applications within the medical field, Gemini-1.5-Pro also performs outstandingly in the understanding of medical images and the analysis of surgical videos. Gemini 1.5 Pro has developed a medical-specific LLM, Med-Gemini-M 1.5~\citep{saab2024capabilities}. This model has been validated in the automated report generation for chest X-rays and CT scans, and some of the results have been rated as "clinically acceptable" by medical experts. In the Direct Video Diagnosis (DVDx) implementation, the system prompt is first provided (details on how it is generated will be discussed later), followed by the user's behavior video as the user prompt. Finally, the model outputs the result.
DVDx requires that MLLM has a powerful video understanding ability, that is, it can "translate" the video content into an understanding of actions and behaviors, and make judgments based on the content understood visually. However, the entire task actually requires MLLM to go through two understanding processes, which is highly likely to lead to the loss or omission of information. Therefore, we innovatively propose HMVDx based on motion understanding and disease diagnosis. This method realizes the process of the MLLM's understanding of the video and the diagnostic process through two models connected in series. For these two models, we conduct fine-tuning of the prompts respectively to achieve their corresponding functions.

HMVDx is developed as a collaborative diagnostic framework based on Gemini-1.5-Pro and DeepSeek-R1, decoupling video description and behavior judgment through the concatenation of multiple models. The role of the Gemini-1.5-Pro model is not to directly produce diagnostic results, but rather to generate detailed descriptions of the movements of individuals in the video. In the diagnostic phase, we employ DeepSeek-R1 as the diagnostic model due to its strong reasoning capabilities~\citep{guo2025deepseek}. Compared to models like GPT-4o, reasoning-oriented models such as DeepSeek-R1 have the advantage of enabling deeper thinking and more advanced reasoning. In addition, as an open-source model, DeepSeek-R1 has a high degree of transparency and lower application costs. In our specific implementation, by leveraging the video understanding ability of Gemini-1.5-Pro, a prompt strategy such as "Please describe the action sequence of the patient in the video using sports medicine terms" is inputted along with the video sample, and the action description is obtained. Taking the action description generated by Gemini-1.5-Pro as the input, DeepSeek-R1 makes judgments by understanding and comparing the description of the human actions with the established rules. Finally, the DeepSeek-R1 model outputs the final diagnostic result.

\subsection{Motion Trajectories Prompt Framework}

The diagnostic prompt is a crucial factor influencing the diagnostic performance of the model~\citep{liu2023pre}. In order to enable the model to fully understand the task requirements, we proposed Motion Trajectories Prompt Framework. In common orthopedic diseases and some neurological diseases, the range of limb movement of patients serves as an important basis for doctors' judgment and understanding of the diseases. During the generation and optimization of the prompts, we propose to use the results of the video understanding by the MLLM as the foundation. At the same time, we replace numerical quantification descriptions with relative position descriptions to improve accuracy, enabling the model to fully understand the situation of human activities and laying the foundation for relevant examinations regarding the trajectory of human movements and the range of limb movements.  

\subsection{Video Understanding-diagnosis Prompt (Prompt-A)}

In the structure of the prompt, we clearly define the role of the model as that of an orthopedic expert to avoid common-sense misjudgments. In the diagnostic thinking path, we guide the model to use the Chain of Thought (COT)~\citep{wei2022chain,zhang2022automatic} method to construct a complete path from action recognition, action judgment to the final result. This conforms to the requirements of the real diagnostic process and ensures the medical compliance of the reasoning path.

Instead of directly using formal medical diagnostic rules, we curated science videos created by doctors from public websites and input them into Gemini-1.5-Pro to generate diagnostic prompts. MLLMs summarized the core actions and diagnostic criteria. We also precisely defined the role of the MLLM in the prompt and elaborated on the requirements during the identification process, such as person confirmation, noise reduction processing, etc. In diagnostic rules, we found that directly using numerical quantitative descriptions (such as "flex the elbow at 30 degrees") would lead to an increase in the misjudgment rate of the model. Through the analysis of the model mechanism, we discovered that the next token prediction mechanism of MLLMs has inherent limitations in understanding the absolute values. Therefore, we switched to using relative position descriptions, such as "higher than the top of the head", etc. In the judgment part about the disease, we constructed a three-dimensional space detection matrix, covering the key movement planes of the shoulder joint. In some detailed designs, the defensive design requires secondary verification for some boundary issues. For example, for key actions (such as touching the back with the hands behind the back), cross-validation prompts are set to prevent single judgment errors; the visual analysis guidance establishes the requirement of \textbf{"watching frame by frame"}, making the model establish the awareness of action trajectory tracking. Finally, we invited medical experts to conduct a secondary confirmation of the action descriptions to ensure that these action descriptions are feasible for the diagnosis and discrimination of shoulder disorders. We took this part of the diagnostic rules as the core component of the prompt.

\subsection{Movement-understanding Prompt (Prompt-B)}
The awareness of the human movement in the video is of vital importance for diagnosis. In this prompt, we established five recognition dimensions, namely movement recognition → spatial trajectory → symmetry comparison → compensation feature → smoothness, which form a complete closed loop for motion analysis.  At the same time, in order to match the diagnostic rules in Prompt-A, which mainly focuses on relative positions, we use a dynamic reference system as well for position description: taking bony landmarks such as the earlobe, acromion, and iliac crest as reference points. This helps the model develop a cognitive understanding of the spatial characteristics of limb movements. For example, interpreting "the height of touching the back reaches the waist" as indicating a lower-than-normal range of motion suggests movement limitation. In addition, for abnormal signals—such as compensatory behaviors like shoulder shrugging—we further enhance the model's ability to recognize and reason about these patterns.

\subsection{Text-judgment Prompt (Prompt-C)}
To align the diagnostic criteria of the two methods, the diagnostic rules in the prompt-C of HMVDx are kept consistent with those in the Direct Video Diagnosis' prompt (prompt - A). Based on the characteristics of the DeepSeek-R1 reasoning model, we guide the model to summarize rule-based movements from the action descriptions, and then judge the possibility of diseases according to the specific performance of movement completion. Meanwhile, we emphasize that identifying potential movement limitations through abnormal signs, such as shoulder shrugging or trembling, is also considered a critical component of the diagnostic process.

\subsection{Evaluation Framework}
Most evaluation metrics in classification are binary ones. They can only enable us to know the final output of the model, but fail to reveal the intermediate output results during the method's execution process. This issue is particularly prominent when MLLMs are applied to the diagnosis of shoulder disorders. 
In the preliminary diagnosis of shoulder disorders, although the final outcome is a binary classification, the reasoning process behind the diagnosis is critically important. To comprehensively evaluate the method, we aim to establish an indicator system that assesses not only the final decision, but also the underlying diagnostic rationale and process. To this end, we propose a comprehensive evaluation framework that includes standard performance metrics—such as accuracy, recall, precision, and F1 score—under various constraints. Additionally, we introduce the Usability Index to further assess the practical applicability of the method.

\subsubsection{Comprehensive metrics}
Accuracy, recall, precision, and F1 score are common evaluation metrics for classification models. 
Recall represents the proportion of actual positive samples that are correctly predicted as positive by the model. Precision refers to the proportion of samples that are actually positive among those predicted as positive by the model. The F1 value is the harmonic mean of precision and recall. It combines precision and recall and can more comprehensively evaluate the performance of the model.

However, it should be noted that in the scenario of judgment of shoulder disorders, traditional classification metrics (such as the F1 value) have limitations. Existing classification metrics only focus on the final judgment and ignore the rationality of the diagnostic path. For example, the model may draw a correct conclusion through incorrect action recognition (such as misjudging "limited forward flexion" as "normal external rotation"). This situation of "correct result but wrong process" poses a high risk in the medical scenario. The rationality and correctness of the medical diagnosis process are equally crucial. An incorrect diagnostic path may lead to misdiagnosis or missed diagnosis, which will have a serious impact on the patient's health. According to the real diagnostic process, doctors need to first identify the patient's actions, then determine whether there are limitations in the actions, and finally draw a conclusion. Therefore, we also incorporate the above judgments into the evaluation indicators, which are divided into whether the action recognition is complete, whether the action judgment is accurate, and whether the final judgment result is correct.

In this research, since the logical consistency of multiple steps is also one of the key aspects of the model's performance that we focus on, when analyzing the classification results output by the model, we have designed a three-level filtering scenario. By gradually strengthening the constraint conditions, we redefine and label the model's output results under different conditions, and quantitatively evaluate the true capabilities of the MLLM method.

\paragraph{Scenario 1: Bottom-line Constraint (Focusing only on the final judgment result)}
In this scenario, the evaluation focuses on the final judgment prediction of the MLLM. That is, the final judgment given by the model is regarded as the prediction result, and traditional evaluation indicators such as F1 values are calculated based on this. This scenario aims to quickly obtain the basic performance of the model at the final judgment level, laying a foundation for subsequent in-depth evaluation.
\paragraph{Scenario 2: Logical Constraint (Taking into account both the final judgment and the movement judgment)}
This scenario focuses on the final judgment of the MLLM on the premise that the behavior judgment is completely reasonable. Only when the model's behavior judgment meets the standard of complete rationality will we accept its final judgment result as the effective prediction of the model. For example, for a positive case of shoulder disorder, if the model's action diagnosis is only partially reasonable, even if its final diagnosis is that the patient has a shoulder disorder, in the evaluation of this scenario, we will still determine the model's final prediction result as 0 (because the model fails to make correct judgments in both movement judgment and final judement, so its final result is not credible). Conversely, if the model's behavior judgment is completely reasonable and the final judgment is that there is a shoulder disorder, then we will determine the model's final prediction result as 1. By introducing the key constraint of the rationality of behavior judgment, this scenario further refines the evaluation of the performance of MLLM, making the evaluation results more in line with the requirements of logical rationality in medical judgment.
\paragraph{Scenario 3: Full-link Constraint (Comprehensively considering the integrity of movement cognition, movement judgment, and the final judgment)}
This scenario is the most stringent in evaluating the MLLM methods, and it is necessary to comprehensively pay attention to the integrity of movement prediction and the rationality of movement judgment. Only when the model achieves complete integrity in movement recognition and the movement judgment is completely reasonable will we refer to its final judgment result. For example, if the model omits key actions in the action prediction link or there are unreasonable aspects in the movement judgment, even if the final judgment result is correct, it will not be regarded as an effective prediction in this scenario. This scenario simulates the strict requirements for the accuracy of the entire process in medical diagnosis, and can comprehensively and deeply reveal the true capabilities of the MLLM methods in complex medical judgment tasks, providing the most comprehensive and strict standard for the accurate evaluation of the performance of the MLLM methods. 

\subsection{Usability Index}

In view of the special requirements for the preliminary diagnosis of shoulder disorders, we have constructed a novel three-dimensional Usability Index system, namely the Usability Index (UI). This system breaks through the limitation of traditional evaluation indicators that only focus on the final result, and achieves a comprehensive and whole-process accurate evaluation of the "judgment path - decision-making logic - result accuracy". Its calculation formula is: UI = 0.5×D + 0.3×R + 0.2×A. In this formula, D represents the accuracy of the final judgment, R represents the rationality of the movement judgment, and A represents the integrity of the movement recognition.

After obtaining the output results of the model for the video samples, the data annotation engineers will carefully break down the conclusions into three sections: movement recognition, movement diagnosis, and final diagnosis, and conduct a comparative analysis with the actual videos to carry out a scoring evaluation. The scoring criteria and their corresponding definitions are presented in Table~\ref{tab:grading_rubric}. In calculating the Usability Index (UI), we involved experienced clinicians to ensure that the evaluation reflects real-world diagnostic reasoning and clinical decision-making. Specifically, the components of Rationality (R) and Accuracy (A) were assessed based on the clinicians' professional judgment. For Rationality (R), clinicians were asked to evaluate whether the model’s decision-making process, such as the sequence of observed signs, the use of spatial reasoning, and the handling of abnormal features, aligned with established medical reasoning and clinical heuristics. For Accuracy (A), the clinicians judged whether the final diagnostic conclusion was consistent with the clinical presentation, using the same standards they would apply in practice. To mitigate subjectivity, we adopted a structured scoring rubric (see Table~\ref{tab:grading_rubric}) and required that each case be independently evaluated by at least two clinicians. In cases of disagreement, a consensus was reached through discussion. This process ensured not only the credibility of the evaluation but also consistency across cases. Involving domain experts in this manner allowed the UI to go beyond conventional metrics (e.g., accuracy or F1 score) and better reflect the method’s practical usability and trustworthiness in clinical scenarios.

Regarding the accuracy of the final judgment D, the rationality of the movement diagnosis R, and the integrity of the movement recognition A, based on the importance and criticality of these three steps in the real medical process, and after discussing with medical experts and collecting the suggestions of multiple experts, we assign weights of 0.5, 0.3, and 0.2 to the three indicators respectively.
\begin{itemize}
    \item $0.5\times$ Accuracy of the Final Judgment (D): In a medical scenario, "accurate judgment results" is the bottom line. Even if there are certain flaws in the judgment process, for example, missing one non-critical action, as long as the final judgment result is correct (D = 1), the output of the model is still feasible.
    \item $0.3\times$ Rationality of the Movement Judgment (R): The weight setting of 0.3 is in a reasonable value, which means that the method is allowed to make certain mistakes in the judgment of some actions, but has to be consistent with the final judgment.
    \item $0.2\times$ Integrity of the Movement Recognition (A): Actions are the foundation for judgment. Missing the detection of key actions (such as failing to recognize "abduction") will lead to the loss of subsequent judgments. The weight of 0.2 neither overly punishes the lack of details (such as missing one non-critical action), nor is it too lenient. 
\end{itemize}
\begin{table}[htbp]
    \centering
    \caption{Comprehensive Metrics}
    \resizebox{\textwidth}{!}{
    \begin{tabular}{cccccc}
    \toprule
    \textbf{Scenario} & \textbf{Framework} & \textbf{Metric} & \textbf{Mean} & \textbf{95\% CI - Lower Bound} &  \textbf{95\% CI - Upper Bound} \\
    \midrule
    \multirow{12}{*}[-10ex]{\parbox{1.8cm}{\centering Scenario 1}}  & \multirow{4}{*}{\parbox{3.5cm}{\centering Baseline}} & Accuracy &	0.489 & 0.349 & 0.628 \\
		& & F1-Score & 0.384 & 0.182 & 0.571 \\
		& & Precision & 0.876 & 0.600 & 1.000 \\
		& & Recall & 0.251 & 0.103 & 0.414 \\
            \cline{2-6}
	& \multirow{4}{*}{\parbox{3.5cm}{\centering Direct Video Diagnosis with Gemini-1.5-Flash}}	& Accuracy & 0.381 & 0.346 & 0.415 \\
		& & F1-Score & 0.151 & 0.110 & 0.192 \\
		& & Precision & 0.824 & 0.714 & 0.929 \\
		& & Recall & 0.083 & 0.059 & 0.108 \\
            \cline{2-6}
	& \multirow{4}{*}{\parbox{3.5cm}{\centering Direct Video Diagnosis with Gemini-1.5-Pro}}	& Accuracy & 0.496 & 0.460 & 0.534 \\
		 &  & F1-Score & 0.409 & 0.363 & 0.460 \\
	 &  & 	Precision & 0.917 & 0.872 & 0.957 \\
	 &  & 	Recall & 0.263 & 0.227 & 0.305 \\
     \cline{2-6}
	 & \multirow{4}{*}{\parbox{3.5cm}{\centering Hybrid Motion Video Diagnosis}} & 	Accuracy	 & 0.883 & 	0.858 & 	0.905 \\
	 &  & 	F1-Score & 0.904 & 0.883 & 0.923 \\
	 &  & 	Precision & 0.988 & 0.976 & 0.998 \\
	 &  & 	Recall & 0.833 & 0.799 & 0.864 \\
     \hline
\multirow{12}{*}[-10ex]{\parbox{1.8cm}{\centering Scenario 2}}	 & \multirow{4}{*}{\parbox{3.5cm}{\centering Baseline}}	 & Accuracy & 	0.370 & 0.233 & 0.512 \\
		 &  & F1-Score	 & 0.120 & 	0.000	 & 0.286 \\
	 &  & 	Precision	 & 0.611 & 	0.000	 & 1.000 \\
	 &  & 	Recall & 	0.068	 & 0.000	 & 0.172 \\
     \cline{2-6}
	 & \multirow{4}{*}{\parbox{3.5cm}{\centering Direct Video Diagnosis with Gemini-1.5-Flash}}	 & Accuracy & 	0.362 & 	0.329 & 	0.398 \\
		 &  & F1-Score	 & 0.133 & 	0.095 & 	0.176 \\
	 &  & 	Precision & 	0.661 & 	0.525	 & 0.786 \\
	 &  & 	Recall & 	0.074	 & 0.052 & 	0.100 \\
     \cline{2-6}
	 & \multirow{4}{*}{\parbox{3.5cm}{\centering Direct Video Diagnosis with Gemini-1.5-Pro}}	 & Accuracy & 	0.381	 & 0.347 & 	0.414 \\
	 &  & 	F1-Score & 	0.177	 & 0.138	 & 0.218 \\
	 &  & 	Precision & 	0.738	 & 0.641 & 	0.836 \\
	 &  & 	Recall & 	0.101 & 	0.077 & 	0.126 \\
     \cline{2-6}
	 & \multirow{4}{*}{\parbox{3.5cm}{\centering Hybrid Motion Video Diagnosis}} & 	Accuracy & 	0.665	 & 0.632	 & 0.696 \\
      &  & F1-Score	 & 0.680	 & 0.641 & 	0.716 \\
	 &  & 	Precision & 	0.926 & 	0.894 & 	0.956 \\
	 &  & 	Recall & 	0.538 & 	0.493 & 	0.581 \\
     \hline
\multirow{12}{*}[-10ex]{\parbox{1.8cm}{\centering Scenario 3}} & \multirow{4}{*}{\parbox{3.5cm}{\centering Baseline}} & 	Accuracy & 	0.023 & 	0.000	 & 0.070 \\
	 &  & 	F1-Score & 	0.045	 & 0.000 & 	0.130 \\
	 &  & 	Precision & 0.063 & 	0.000 & 	0.214 \\
	 &  & 	Recall & 	0.036 & 	0.000 & 	0.120 \\
     \cline{2-6}
	 & \multirow{4}{*}{\parbox{3.5cm}{\centering Direct Video Diagnosis with Gemini-1.5-Flash}}	 & Accuracy & 	0.175 & 	0.147 & 	0.204 \\
	 &  & 	F1-Score & 	0.090 & 	0.060 & 	0.122 \\
	 &  & 	Precision & 	0.167 & 	0.112 & 	0.224 \\
	 &  & 	Recall & 	0.061 & 	0.040 & 	0.085 \\
     \cline{2-6}
	 & \multirow{4}{*}{\parbox{3.5cm}{\centering Direct Video Diagnosis with Gemini-1.5-Pro}}	 & Accuracy	 & 0.228 & 	0.197	 & 0.258 \\
	 &  & 	F1-Score & 	0.098 & 	0.069 & 	0.130 \\
	 &  & 	Precision	 & 0.218 & 	0.154 & 	0.283 \\
	 &  & 	Recall & 	0.064	 & 0.044 & 	0.085 \\
     \cline{2-6}
	 & \multirow{4}{*}{\parbox{3.5cm}{\centering Hybrid Motion Video Diagnosis}} & Accuracy & 	0.272	 & 0.242	 & 0.306 \\
		 &  & F1-Score & 	0.194 & 	0.153 & 	0.237 \\
	 &  & 	Precision & 	0.363 & 	0.291 & 	0.434 \\
	 &  & 	Recall & 	0.133 & 	0.103	 & 0.165 \\
    \bottomrule
    \end{tabular}
    }
    \label{tab:comprehensive_matrix}
\end{table}

\newpage

\bibliography{sample}
\bibliographystyle{apalike}

\end{document}